\pdfoutput=1
\documentclass[11pt]{article}

\usepackage[final]{acl}

\usepackage{times}
\usepackage{latexsym}

\usepackage[T1]{fontenc}
\usepackage[utf8]{inputenc}

\usepackage{microtype}

\usepackage{inconsolata}

\usepackage{graphicx}

\usepackage{soul}
\usepackage{url}
\usepackage{graphicx}
\usepackage{amsmath}
\usepackage{amsthm}
\usepackage{booktabs}
\usepackage{algorithm}
\usepackage{algorithmic}
\usepackage{multirow}
\usepackage{wrapfig}
\usepackage{arydshln}
\usepackage{multirow} 
\usepackage{enumerate}
\usepackage{colortbl}

\title{Invoke Interfaces Only When Needed: Adaptive Invocation for Large Language Models in Question Answering}

\author{
  \,Jihao Zhao\textsuperscript{1} \quad
  Chunlai Zhou\textsuperscript{1} \quad
  Daixuan Li\textsuperscript{1} \quad
  Shuaishuai Zu\textsuperscript{1} \quad
  Biao Qin\textsuperscript{1}\thanks{Corresponding author: \texttt{qinbiao@ruc.edu.cn}} \\
  \,\textsuperscript{1}School of Information, Renmin University of China, Beijing, China 
}

\begin{document}
\maketitle
\begin{abstract}
The collaborative paradigm of large and small language models (LMs) effectively balances performance and cost, yet its pivotal challenge lies in precisely pinpointing the moment of invocation when hallucinations arise in small LMs. Previous optimization efforts primarily focused on post-processing techniques, which were separate from the reasoning process of LMs, resulting in high computational costs and limited effectiveness. In this paper, we propose a practical invocation evaluation metric called AttenHScore, which calculates the accumulation and propagation of hallucinations during the generation process of small LMs, continuously amplifying potential reasoning errors. By dynamically adjusting the detection threshold, we achieve more accurate real-time invocation of large LMs. Additionally, considering the limited reasoning capacity of small LMs, we leverage uncertainty-aware knowledge reorganization to assist them better capture critical information from different text chunks. Extensive experiments reveal that our AttenHScore outperforms most baselines in enhancing real-time hallucination detection capabilities across multiple QA datasets, especially when addressing complex queries. Moreover, our strategies eliminate the need for additional model training and display flexibility in adapting to various transformer-based LMs. Our code is available at \url{https://github.com/Robot2050/AttenHScore}.
\end{abstract}

\begin{figure*}[t]
    \centering
    \includegraphics[width=\textwidth]{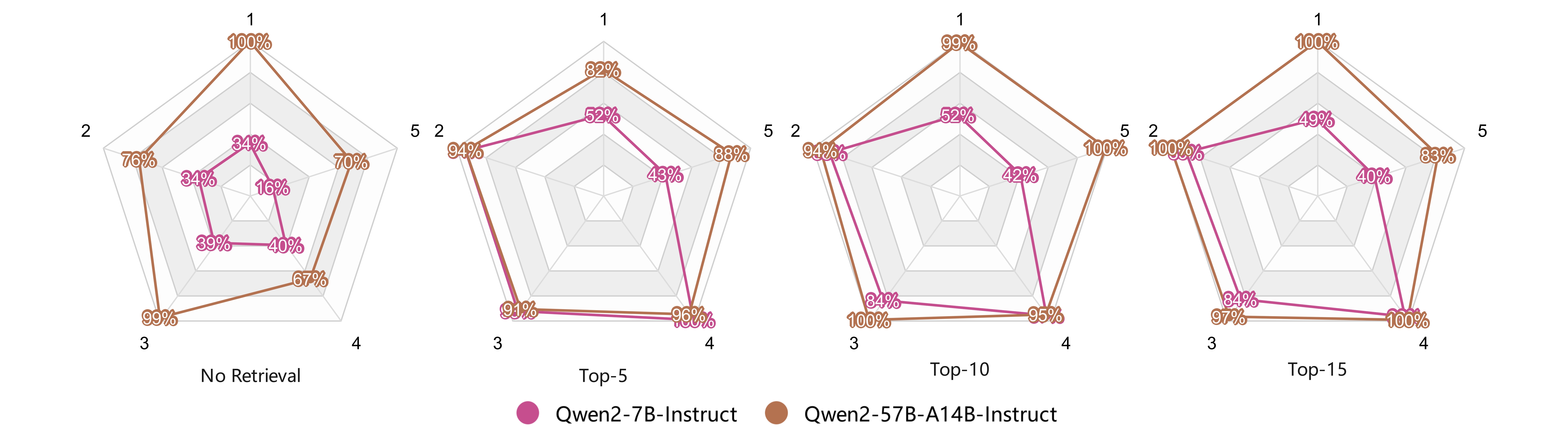}
    \caption{Performance of large and small LMs on different QA datasets in the RAG scenario.  Including 1: 2WikiMultihopQA, 2: MultiFieldQA-en, 3: Qasper, 4: MultiFieldQA-zh and 5: HotpotQA.}
    \label{fig:leidatu}
\end{figure*}

\section{Introduction}
With the profound study of the scaling law \cite{kaplan2020scaling} and the density law \cite{xiao2024densing}, the development and application of language models (LMs) have exhibited a diversified pattern. In this context, the remarkable performance of large language models (LLMs) such as GPT-4o in reasoning tasks has attracted significant attention \cite{hosseini2023exploratory}. However, due to their complex structures and massive parameter scales, these LLMs consume considerable computational resources during training and inference. Consequently, many of these LLMs are only available through paid API services, undoubtedly increasing their monetary cost. Meanwhile, small language models (SLMs), with their lightweight architectures and efficient inference capabilities \cite{zhang2024tinyllama}, demonstrate significant advantages in specific scenarios, such as real-time responses on edge devices \cite{khiabani2025optimizing} and rapid processing of simple tasks \cite{li2024purifying}. Nevertheless, when faced with higher-level tasks requiring complex semantic understanding, the capabilities of SLMs appear to be inferior compared to those of LLMs \cite{wang2024comprehensive}.

To balance performance and cost while enhancing overall efficiency, a new paradigm of collaboration between large and small LMs has emerged from the perspectives of cost-effectiveness and resource optimization. This paradigm aims to fully leverage the advantages of LLMs in handling complex tasks while exploiting the efficiency of SLMs in simple problem scenarios, thus achieving optimal resource allocation and efficient task processing. As illustrated in Figure \ref{fig:leidatu}, we conduct retrieval-based question answering (QA) experiments utilizing two LMs, one large and one small, across five datasets from Longbench \cite{bai2023longbench}, to evaluate the performance of both LMs in scenarios without retrieval, and with top-5, top-10, top-15 retrieval results. LLM exhibits overall superior performance, but the gap between it and SLM is remarkably narrow on certain datasets. Under these circumstances, researchers have mainly proposed two strategies: routing and cascading. The core mechanism of the former lies in accurately directing user queries to a specific model based on criteria provided by specially trained models \cite{aggarwal2023automix,ding2024hybrid}. Comparatively, the latter exhibits a more flexible and phased processing mode. According to this strategy, user queries are first sent to SLMs for initial processing. Then, based on the output results of these models, the system determines whether further in-depth reasoning by LLMs is necessary \cite{yue2023large,ramirez2024optimising}.

Based on the aforementioned research, we find that routing strategies require the introduction of auxiliary models for decision-making during implementation, which contradicts the initial goal of simplicity and efficiency. More importantly, these auxiliary models not only require specialized training but also often rely on specific datasets \cite{vsakota2024fly,ding2024hybrid}, potentially limiting their versatility across different tasks. In view of this, we have chosen to adopt a cascading strategy, where the main technical challenge lies in accurately determining when hallucinations occur in SLMs. Currently, research on hallucination detection in LMs primarily focuses on the post-reasoning phase \cite{manakul2023selfcheckgpt,zhang2023enhancing,li2023halueval}. However, such methods exhibit significant limitations when integrated into the practical LLMs applications. The primary issue is that these post-processing methods often incur high computational costs and notable delays. For instance, cutting-edge detection methods typically utilize LLMs such as ChatGPT, OPT, etc. \cite{zhang2023enhancing}, making the cost of hallucination detection comparable to or even more expensive than LLMs reasoning tasks. What's more, post-processing methods are independent of the reasoning process \cite{shi2022natural,wang2022self}, thus they cannot delve into the origins and evolution of hallucinations within each LMs.

Seeking to surmount the outlined restrictions, we shift the focus of optimizing LMs invocations towards understanding their existing available signals, rather than training and running more auxiliary models. This paper proposes a practical invocation evaluation metric, AttenHScore, designed to calculate the accumulation and propagation of hallucinations during the generation process of SLMs. By continuously amplifying potential error points, this metric enables more skillfully identify deviations between generated content and facts, thereby improving the detection accuracy of hallucinations. Furthermore, from the perspective of retrieval-augmented generation (RAG), we guide SLMs to evaluate the uncertainty between queries and different text chunks, optimizing the information arrangement by moving more relevant content from the retrieval to the front of the prompt, thereby further assisting SLMs in capturing key information and enhancing their accuracy in QA tasks.

The main contributions of this work are as follows:
\begin{itemize}
    \item We propose a method for optimizing LMs invocation based on the uncertainty of generated text. The core technology lies in the thorough consideration of accumulation and propagation effects of hallucinations, thereby achieving unsupervised, real-time and plug-and-play invocation optimization.
    \item In the realm of retrieval-based QA, we fully utilize the chain-of-thought reasoning capability of generative LMs and guide text re-ranking through an uncertainty evaluation mechanism to precisely optimize information arrangement.
    \item To validate the effectiveness of our method, we test it on four QA datasets utilizing three different LLMs and conduct an in-depth analysis of the proposed method through multi-dimensional experiments.
\end{itemize}

\section{Related Works}

\textbf{Collaboration of SLMs and LLMs} \quad The joint application of LLMs and SLMs has recently emerged as a technological approach, achieving breakthroughs in multiple research areas \cite{ma2023large,ding2024rationale,min2024synergetic}. In the studies by Sakota et al. \shortcite{vsakota2024fly} and Lu et al. \shortcite{lu2023routing}, they proposed training an auxiliary model to estimate the success rate of invoking LLMs. Chen et al. \shortcite{chen2023frugalgpt} introduced a cascade strategy, utilizing an auxiliary model to predict the accuracy of outputs from SLMs. Additionally, Yue et al. \shortcite{yue2023large} suggested repeatedly invoking SLMs to perform inference tasks, while research by Ram{\'\i}rez et al. \shortcite{ramirez2023cache} indicated that the margin of a knowledge-distilled model has the potential to enhance the efficiency of calls made to LLMs. Later, Ram{\'\i}rez et al. \shortcite{ramirez2024optimising} proposed the Margin Sampling approach, which identifies hallucinations by computing the margin between the most likely first and second tokens. However, the above methods are more suitable for short answer generation tasks, while the direct judgment of long answer generation is still a gap and more challenging.

\textbf{Hallucination Detection} \quad The concept of hallucination, which originally emerged from the fields of pathology and psychology \cite{macpherson2013philosophy}, has been subsequently adopted and applied in the domain of Natural Language Processing \cite{maynez2020faithfulness}. The occurrence of hallucinations is widespread in deep learning models utilized for a range of text generation tasks \cite{dziri2022origin,su2022read}. It is defined as the generation of content that lacks practical significance or deviates from the provided source material \cite{ji2023survey}. With the widespread adoption of LLMs in various applications, the issue of hallucinations arising from these LMs has garnered significant attention from researchers \cite{shen2023chatgpt,becker2024text}. In this context, Min et al. \shortcite{min2023factscore} introduced the FactScore method, which leverages knowledge sources to verify the accuracy of each atomic fact in the generated text. Furthermore, Manakul et al. \shortcite{manakul2023selfcheckgpt} presented SelfCheckGPT in their study, a black-box technique for hallucination detection. Despite those advancements, their methods still possess certain limitations. They either rely on external knowledge bases or require the analysis of multiple responses sampled from LMs, which undoubtedly increases resource consumption and reduces efficiency.

\section{Optimizing the Adaptive Invocation Interface for LLMs}
Our design philosophy is to "maximize the success rate of SLMs and minimize the need for LLM invocations", thereby embodying the idea of "Invoke Interfaces Only When Needed". We achieve this goal through a two-stage strategy that combines "pre-event optimization" with "post-event quality control":

Re-ranking: This module aims to reduce the likelihood of SLMs generating hallucinations at the source. As we pointed out in the paper, SLMs have limited capabilities in processing long texts and extracting key information. By re-ranking the input information and placing the most critical content at the forefront, we can significantly optimize the working environment for SLMs, thereby proactively minimizing scenarios that might necessitate LLM invocations.

Hallucination Detection: When "pre-event optimization" is still insufficient to completely avoid errors, this module serves as a safety net. It is responsible for real-time monitoring of SLM outputs to precisely identify moments when LLM invocations are truly necessary.

\subsection{Problem Definition}
In this paper, we focus on predicting the mapping relationship between elements in the input space $X$ and their corresponding labels in the output space $Y$. Here, $(x_1, ..., x_q) \sim X$ represents the response generated by SLMs upon a user query, while $(0, 1) \sim Y$ denotes the decision flag indicating whether to invoke LLMs. We transform the system into the predictor $f: X \rightarrow Y$. For each incoming $X$, we determine whether to call LLMs based on the hallucination detection strategy. The entire procedure is outlined in Algorithm \ref{alg1}.

\subsection{Real-time Hallucination Detection}
Our work is grounded in a key insight: hallucinations are not instantaneous and isolated errors but rather a conductive and gradually accumulating dynamic process. The uncertainty signals from individual tokens are often weak and ambiguous. However, when these signals appear continuously and are amplified along the inference chain, they constitute a strong warning that hallucinations are about to occur.

\begin{figure}[t]
    \centering
    \includegraphics[width=0.47\textwidth]{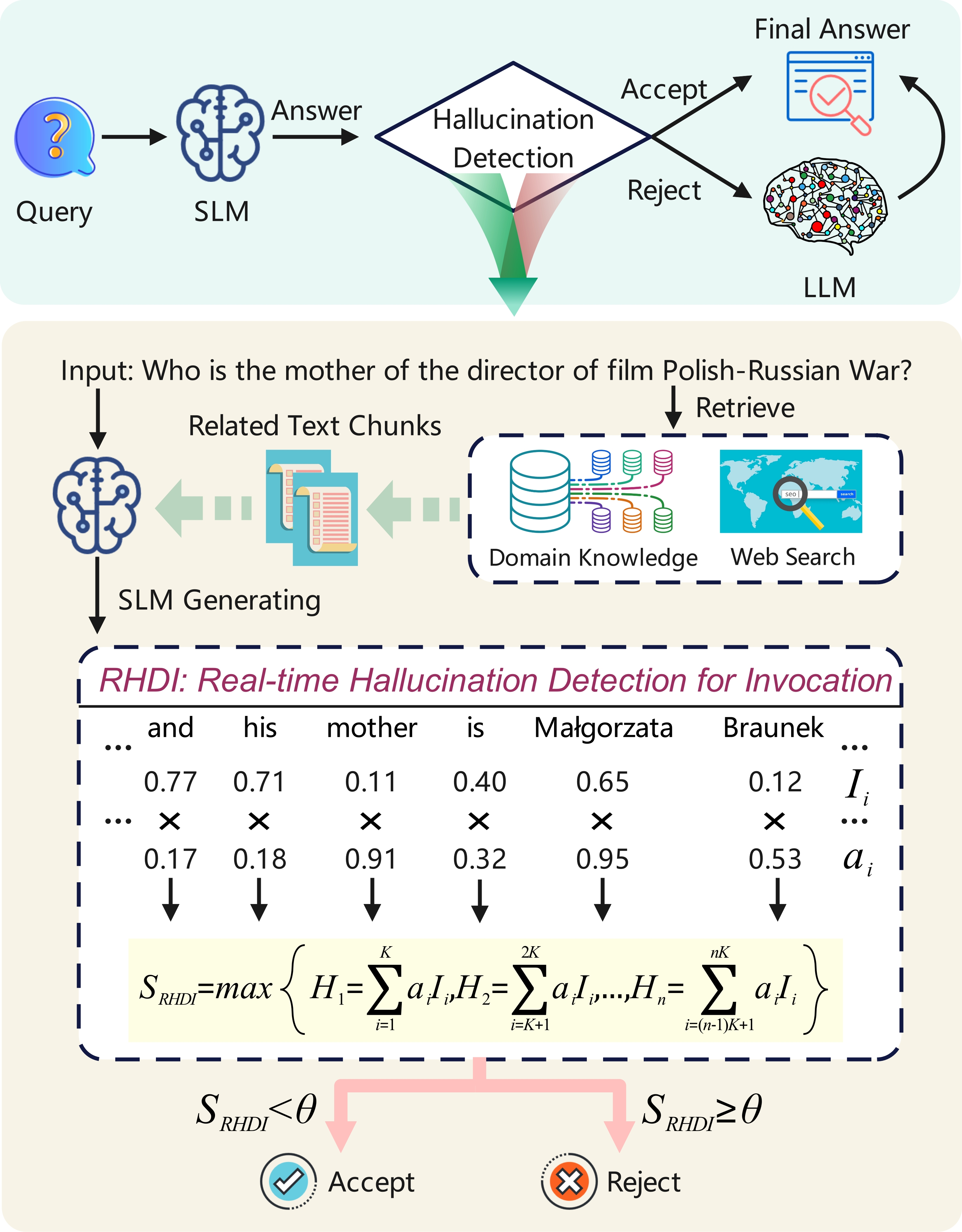}
    \caption{Overview of our hallucination detection and collaborative framework.}
    \label{fig:kuangjia}
\end{figure}

Based on the aforementioned in-depth analysis, we ascertain that current methods relying on post-processing or uncertainty measures are inadequate for detecting hallucinations in collaborative large and small LMs systems. Given this limitation, we address the issue from the perspective of sequence generation in SLMs. Observing the accumulation and propagation of hallucinations during the token-by-token generation process, we propose the AttenHScore evaluation metric to quantify these characteristic, thereby providing valuable guidance for hallucination detection. As illustrated in Figure \ref{fig:kuangjia}, we define this metric as follows:
\begin{eqnarray}
H=\sum_{i=1}^{K}a_{i}I_{i}=-\sum_{i=1}^{K}a_i\log p_{max}(x_i)
\end{eqnarray}
where $p_{max}(x_i)$ represents the maximum probability of generating token $x_i$ at position $i$, $I_i$ denotes the degree of uncertainty for that token, and $a_i$ signifies the accumulation and propagation weight of hallucination designed for each $I_i$, which is specifically calculated as:
\begin{eqnarray}
a_i= p_{max}(x_i)\mathrm{Atten}(x_i)
\end{eqnarray}
Specifically, $\mathrm{Atten}(x_i)$ is used in attention-based models to measure the degree of attention the LM pays to each token, reflecting which tokens are more important and relevant for answering in the current processing step. By multiplying $p_{max}(x_i)$ and the attention score, we obtain a weight that comprehensively reflects the degree of attention and confidence of the token during model processing. Therefore, the above two steps of accumulation and multiplication together highlight the hallucinations of LMs during generation more effectively.

If the generated text is long, we preset a value $K$, calculate an AttenHScore value for every $K$ tokens, and take the maximum as the object to compare with the threshold to determine whether to invoke the LLM:
\begin{eqnarray}
S_{RHDI}=\max\left\{H_1,H_2,...,H_n\right\}
\end{eqnarray}

In addition, we conduct a comprehensive design for the computation of $\mathrm{Atten}(x_i)$. Initially, we integrate the softmax function with the mask function to generate the attention weight matrix $M$:
\begin{eqnarray}
M=\text{softmax}\left(\text{mask}\left(\frac{QR^{\top}}{\sqrt{d_{k}} } \right)\right)
\end{eqnarray}
where $Q$ represents the query matrix, $R$ stands for the key matrix, and $d_k$ denotes the dimensionality of a key vector. Following this, for a given token $x_i$, we determine its corresponding maximum attention value by searching through all elements $M_{j,i}$ where $j > i$. Lastly, to further enhance the influence of attention scores in the overall evaluation, we employ an exponential function to amplify them:
\begin{eqnarray}\label{eq4}
\mathrm{Atten}(x_i)=\exp\left(\max_{j > i} M_{j,i}\right)
\end{eqnarray}
Through this approach, we ensure that the attention mechanism plays a more prominent role in the evaluation. It is worth noting that in Eq. (\ref{eq4}), we choose max to calculate attention scores, rather than basing on a specific layer or taking an average. This is because we believe that doing so may be affected by special cases and reduce the perception of key content, thereby affecting our final detection performance. This is experimentally confirmed in Section \ref{Ablation studies}.

\subsection{Dynamic Threshold}
Setting a threshold for decision criteria is a common requirement across all strategies, and we introduce a dynamic threshold mechanism. Specifically, we first utilize the results of the first five queries to calculate an initial threshold. During this process, we do not evaluate whether these five queries trigger the LLM, but only obtain output results from the SLM. Subsequently, at each new query, we incorporate the hallucination score of the current query into the historical records and recalculate the average hallucination score of all processed queries, using this as the updated threshold. 

\subsection{Re-ranking Strategy based on Uncertainty Evaluation}
In long text processing scenarios, SLMs often face challenges in extracting effective information, leading to inefficient utilization of key information. Additionally, these SLMs exhibit a significant position bias phenomenon in long texts, where they tend to focus more on the beginning of the prompt and easily overlook information in the middle \cite{jiang2023longllmlingua}. Therefore, we introduce auxiliary mechanisms for SLMs to enhance their information utilization capabilities.

Given a query, we are able to retrieve multiple associated text chunks. For each text chunk, we guide SLMs to perform reverse thinking, which involves generating the corresponding query based on the text content. Afterwards, we quantify the uncertainty of this generation process using the following method:
\begin{eqnarray}
G=-\sum_{x_i\in X }\mathrm{Atten}(x_i)\log p(x_i)
\end{eqnarray}
where $X$ represents the token set of the known query. This approach takes full advantage of the powerful reasoning capabilities and deep understanding of structural nuances inherent in current LMs. Experimental results presented in Section \ref{Re-ranking} suggest that this method possesses generalization capabilities, enabling it to more accurately filter out noisy or incomplete information when compared to prevailing benchmark models.

By integrating the various strategies we proposed, real-time hallucination detection and re-ranking  are achieved within a large and small LMs collaboration system without the need for additional model training. This process is unsupervised, namely, our methods do not require manual supervision or labeled data for training. More meaningfully, our methods are universally applicable to all transformer-based LMs, truly embodying the plug-and-play principle and showcasing flexibility.

\begin{algorithm} 
    \renewcommand{\algorithmicrequire}{\textbf{Input:}}
    \renewcommand{\algorithmicensure}{\textbf{Output:}}
    \caption{Adaptive Invocation for LLMs in QA} 
    \label{alg1} 
    \begin{algorithmic}[1]
        \REQUIRE SLM generator $M_s$, LLM interface $M_l$, User query $Q_i$, 
    Initial threshold $\theta$
        \ENSURE Decision $y \in \{0,1\}$ for LLM invocation, Response $R$
        \WHILE{new user query $Q_i$ arrives} 
        \STATE $M_s(Q_i)$ generate candidate tokens $X = \{x_1,...,x_q\} \rightarrow  \text{logits}, \text{attentions}$
        \IF{$i \le  5$} 
        \STATE $y \gets 0, R(Q_i) \gets X$ 
        \ELSE 
        \STATE Calculate $\mathrm{Atten}(x_i)$
        \STATE Gradually calculate $H_1,H_2,...,H_n$
        \STATE $S_{RHDI} \gets \max\left\{H_1,H_2,...,H_n\right\}$
        \IF{$S_{RHDI}< \theta$} 
        \STATE $y \gets 0, R(Q_i) \gets X$  
        \ELSE 
        \STATE $y \gets 1, R(Q_i) \gets M_l(Q_i)$ 
        \ENDIF
        \STATE Upadte $\theta \gets \frac{\sum_{k=1}^{N} S_{RHDI}(X_k)}{n}$
        \ENDIF
        \ENDWHILE
    \end{algorithmic} 
\end{algorithm}

\renewcommand{\arraystretch}{1.2} 
\setlength{\extrarowheight}{1pt} 
\begin{table*}[h]
\centering
\resizebox{\textwidth}{!}{%
\begin{tabular}{lccc|ccc|ccc|ccc}
\toprule
\textbf{Dataset} & \multicolumn{3}{c}{\textbf{CoQA}} & \multicolumn{3}{c}{\textbf{SQuAD}} & \multicolumn{3}{c}{\textbf{TriviaQA}} & \multicolumn{3}{c}{\textbf{NQ}}  \\
\textbf{Method} & \textbf{AUCs} & \textbf{AUCr} & \textbf{ACCr} & \textbf{AUCs} & \textbf{AUCr} & \textbf{ACCr} & \textbf{AUCs} & \textbf{AUCr} & \textbf{ACCr} & \textbf{AUCs} & \textbf{AUCr} & \textbf{ACCr} \\
\midrule
\multicolumn{13}{c}{\textit{Llama3-8B-Instruct}} \\ 
\addlinespace[2pt] % Add space after italicized line
\cdashline{1-13} % Dashed line after italicized row
Perplexity& 0.5783 & 0.5509 & 0.5251 & 0.4745 & 0.4645 & 0.4782 & \textbf{0.8431} & \underline{0.8342} & \textbf{0.7525} & \textbf{0.7700} & \underline{0.7694} & 0.6742 \\
Energy& 0.4212 & 0.3797 & 0.4025 & 0.4297 & 0.4129 & 0.4497 & 0.7204 & 0.6920 & 0.6547 & 0.6551 & 0.6440 & 0.6111 \\
AVG-Range& 0.5344 & 0.5033 & 0.5068 & 0.4609 & 0.4516 & 0.4821 & 0.8277 & 0.8229 & 0.7473 & 0.7492 & 0.7555 & 0.7172 \\
LN-Entropy& 0.6732 & 0.6668 & 0.6280 & 0.6113 & 0.6134 & 0.6179 & 0.8189 & 0.8177 & \underline{0.7518} & 0.7490 & 0.7553 & 0.6640 \\
LexicalSimilarity& 0.7602 & 0.7614 & 0.7041 & 0.6365 & 0.6341 & 0.5562 & 0.7838 & 0.7746 & 0.7412 & 0.7354 & 0.7321 & \textbf{0.7172} \\
EigenScore& \underline{0.7910} & \underline{0.8014} & \underline{0.7328} & \underline{0.7359} & \underline{0.7417} & \underline{0.6741} & 0.7941 & 0.7783 & 0.7410 & 0.7599 & 0.7587 & 0.6801 \\
\rowcolor[rgb]{0.85, 0.92, 0.99} AttenHScore (Ours)& \textbf{0.8330} & \textbf{0.8706} & \textbf{0.8097} & \textbf{0.8715} & \textbf{0.9024} & \textbf{0.8176} & \underline{0.8334} & \textbf{0.8388} & 0.7513 & \underline{0.7650} & \textbf{0.7871} & \underline{0.7072} \\
\midrule
\multicolumn{13}{c}{\textit{Vicuna1.5-7B}} \\ 
\addlinespace[2pt] % Add space after italicized line
\cdashline{1-13} % Dashed line after italicized row
Perplexity& 0.4701 & 0.3292 & 0.3492 & 0.5143 & 0.2610 & 0.3109 & \textbf{0.8184} & \underline{0.8108} & 0.7366 & 0.6794 & 0.6794 & 0.6427 \\
Energy& 0.3817 & 0.2139 & 0.2307 & 0.4273 & 0.1648 & 0.1791 & 0.7316 & 0.7147 & 0.6632 & 0.5767 & 0.5613 & 0.4947 \\
AVG-Range& 0.4624 & 0.3128 & 0.4154 & 0.5164 & 0.2645 & 0.3591 & 0.7859 & 0.7820 & 0.7165 & 0.6395 & 0.6344 & 0.6615 \\
LN-Entropy& 0.5221 & 0.4274 & 0.3739 & 0.5672 & 0.4331 & 0.5383 & 0.7962 & 0.7974 & 0.7339 & 0.6792 & 0.6895 & 0.6593 \\
LexicalSimilarity& 0.5876 & 0.5518 & 0.4894 & 0.5656 & 0.4650 & 0.5530 & 0.7870 & 0.7833 & 0.7385 & 0.7279 & 0.7441 & \textbf{0.7443} \\
EigenScore& \underline{0.6500} & \underline{0.6648} & \underline{0.5165} & \underline{0.6441} & \underline{0.6315} & \underline{0.5309} & 0.7979 & 0.7880 & 0.7402 & \textbf{0.7557} & \underline{0.7748} & 0.6825 \\
\rowcolor[rgb]{0.85, 0.92, 0.99} AttenHScore (Ours)& \textbf{0.7503} & \textbf{0.8481} & \textbf{0.7840} & \textbf{0.7193} & \textbf{0.8085} & \textbf{0.7212} & \underline{0.8178} & \textbf{0.8338} & \textbf{0.7467} & \underline{0.7524} & \textbf{0.7949} & \underline{0.6958} \\
\midrule			
\multicolumn{13}{c}{\textit{Llama2-13B-Chat-HF}} \\ 
\addlinespace[2pt] % Add space after italicized line
\cdashline{1-13} % Dashed line after italicized row
Perplexity& 0.5423 & 0.5272 & 0.5108 & 0.4830 & 0.4638 & 0.4504 & \textbf{0.8111} & \underline{0.8142} & \underline{0.7422} & 0.6944 & 0.6942 & 0.6463 \\
Energy& 0.4380 & 0.3993 & 0.4596 & 0.4102 & 0.3890 & 0.4167 & 0.6976 & 0.6888 & 0.6545 & 0.6229 & 0.6133 & 0.5507 \\
AVG-Range& 0.5243 & 0.5075 & 0.5569 & 0.4651 & 0.4451 & 0.4612 & 0.7936 & 0.8002 & 0.7276 & 0.6570 & 0.6562 & 0.6620 \\
LN-Entropy& 0.6005 & 0.6018 & 0.5867 & 0.5938 & 0.5904 & 0.5778 & 0.7729 & 0.7855 & 0.7208 & 0.6849 & 0.6931 & 0.6169 \\
LexicalSimilarity& 0.7155 & 0.7331 & 0.6593 & 0.6536 & 0.6667 & 0.6623 & 0.7439 & 0.7466 & 0.7303 & 0.7286 & 0.7373 & \underline{0.6928} \\
EigenScore& \underline{0.7509} & \underline{0.7809} & \underline{0.7120} & \underline{0.7364} & \underline{0.7585} & \underline{0.6670} & 0.7512 & 0.7502 & 0.7265 & \textbf{0.7477} & \underline{0.7645} & 0.6717 \\
\rowcolor[rgb]{0.85, 0.92, 0.99} AttenHScore (Ours)& \textbf{0.8369} & \textbf{0.8982} & \textbf{0.8320} & \textbf{0.8544} & \textbf{0.9032} & \textbf{0.8322} & \underline{0.8036} & \textbf{0.8221} & \textbf{0.7442} & \underline{0.7423} & \textbf{0.7785} & \textbf{0.6978} \\
\bottomrule
\end{tabular}%
}
\caption{Main experimental results are presented in four QA datasets. The best result is in bold, and the second best result is underlined.}
\label{tab:main-performance}
\end{table*}

\section{Experiment}

\subsection{Datasets and Metrics}
We adopt four highly recognized QA datasets for evaluation, including two open-book conversational datasets: CoQA \cite{reddy2019coqa} and SQuAD \cite{rajpurkar2016squad}, and two closed-book QA datasets: TriviaQA \cite{joshi2017triviaqa} and Natural Questions (NQ) \cite{kwiatkowski2019natural}. CoQA is sourced from seven different domains, with each dialogue involving two crowd workers engaging in a question-and-answer exchange around a passage \cite{reddy2019coqa}. SQuAD is renowned for its large scale and high quality, with its origins in Wikipedia articles \cite{rajpurkar2016squad}. TriviaQA is a reading comprehension dataset that comprises question-answer-evidence triplets \cite{joshi2017triviaqa}. NQ contains authentic queries posed by users to Google Search, along with answers sourced from Wikipedia \cite{kwiatkowski2019natural}. On the other hand, the actual answers in the CoQA and SQuAD datasets are often longer, whereas answers in the TriviaQA and NQ datasets tend to be in the form of single or few-word responses. For evaluation metrics, we follow the prior work of Ren et al. \shortcite{ren2022out} and Chen et al. \shortcite{chen2024inside} by utilizing the area under the receiver operator characteristic curve (AUROC) and accuracy (ACC). Specifically, AUCs denotes the AUROC score with sentence similarity serving as the measure of correctness, while AUCr represents the AUROC score with the Rouge-L score as the correctness measure, and ACCr follows similarly.

\renewcommand{\arraystretch}{1.2} 
\setlength{\extrarowheight}{1pt} 
\begin{table*}[ht]
\centering
\resizebox{\textwidth}{!}{%
\begin{tabular}{lccccccccc}
\toprule
 \textbf{Methods} & \textbf{ERNIE-3.5} & \textbf{Qwen-Plus} & \textbf{Qwen-Turbo} & \textbf{Deepseek-v3} & \textbf{Qwen-72B} & \textbf{Qwen1.5-72B} & \textbf{Qwen2-57B} & \textbf{R1-LLama-70B} & \textbf{R1-Qwen-32B}   \\
\midrule
\multicolumn{10}{l}{\textit{Initial Score with Only Vicuna-7B-v1.5: 13.25; Score After Our Re-ranking Process: 16.62.}} \\
\midrule
\multicolumn{10}{l}{\textit{With Large-Small Language Model Collaboration}} \\
\addlinespace[2pt] % Add space after italicized line
\cdashline{1-10} % Dashed line after italicized row
Perplexity& 17.82&17.01&20.31&17.27&18.76&19.14&18.02&17.51&17.69\\
Random & 21.05 & 18.9 & 22.61 & 19.53 & 20.89 & 22.16 & 20.28 & 20.72 & 20.28 \\
AVG\_Range & 22.33 & 20.87 & 23.49 & 20.39 & 21.93 & 22.59 & 21.62 & 21.65 & 21.34 \\
\rowcolor[rgb]{0.85, 0.92, 0.99} AttenHScore& \textbf{24.91} & \textbf{23.27} &  \textbf{26.23} & \textbf{22.82} & \textbf{24.95} & \textbf{25.71} & \textbf{23.69} & \textbf{23.76} & \textbf{24.03}  \\
\midrule
\multicolumn{10}{l}{For Reference: \textit{Scores Obtained by Exclusively Utilizing Interfaces of Various Large Language Models.}} \\
\addlinespace[2pt] % Add space after italicized line
\cdashline{1-10} % Dashed line after italicized row
LLMs & 25.12 & 22.25 & 27.71 & 22.0 & 26.05 & 27.45 & 23.9 & 23.65 & 23.5 \\
\bottomrule
\end{tabular}%
}
\caption{We report the metric F1 score of QA performance under three scenarios: SLM only, large-small LM collaboration, and LLM only.}
\label{tab:allllm}
\end{table*}

\subsection{Baselines}
We undertake a comparative analysis of our proposed approach with the prevalent uncertainty-based techniques, namely Length-normalized Entropy (LN-Entropy) \cite{malinin2020uncertainty}, the consistency-based metric Lexical Similarity \cite{lin2022towards} as well as EigenScore \cite{chen2024inside}, which utilizes the eigenvalues of the response covariance matrix to quantify semantic consistency or diversity in the dense embedding space. All three aforementioned methods require SLMs to generate multiple answers to the same question. In addition, we introduce three comparison methods that only require SLMs to generate an answer once. Perplexity evaluates the rationality of text generation by calculating the predictive probability distribution of SLMs \cite{ren2022out}. AVG-Range assesses credibility by measuring the average difference between the highest and lowest probabilities in the probability distribution of each token output by SLMs \cite{ramirez2024optimising}. Energy score \cite{liu2020energy}, a popular out-of-distribution detection method, is tested for its applicability in hallucination detection. Our methodology also adheres to the paradigm of single-pass model generation.
% Where SLMs perform the generation process only once within the large-and-small model collaboration framework.

\subsection{Implementation Setting}
In experiments aimed at detecting hallucinations for collaboration, we primarily employ three LMs with the following hyperparameter settings: temperature at 0.5, top-p at 0.99, top-k at 5, and the number of generations set to 10. When assessing the correctness of generated answers, we adopt two commonly used methods: Rouge-L \cite{lin2004rouge} and semantic similarity \cite{reimers2019sentence}. The former employs the threshold of 0.5, while the latter utilizes the nli-roberta-large model with the threshold set to 0.9. 
% When utilizing the uncertainty derived from SLMs to facilitate re-ranking, we opt for two SLMs, retaining their default hyperparameter configurations. 
Moreover, in conducting collaborative experiments between small and large LMs, we select Vicuna-7B-v1.5 as the SLM and incorporated nine distinct LLM interfaces to participate in the experiments. We incorporate RAG techniques, using bge-large-en-v1.5 as the retriever and setting the number of retrieved text chunks to 10. Detailed experimental setup information can be found in Appendix \ref{sec:appendix2}.

\subsection{Main Results}
In this section, we first conduct a comprehensive evaluation of the key component for detecting hallucinations in SLMs within the collaborative system of large-small LM on the hallucination benchmark \cite{chen2024inside}. Subsequently, we integrate AttenHScore into the entire system and evaluate its accuracy in determining interface calls by comparing various real-time hallucination detection methods.

\subsubsection{Overall Results of the Hallucination Detection Component}
To comprehensively validate the effectiveness of our proposed AttenHScore, we conduct experiments exploiting three LMs and four widely-used QA datasets. In designing the experiments, we not only consider the diversity of baseline methods but also emphasize the comprehensiveness of evaluation metrics to ensure the objectivity and accuracy of assessment results. The experimental results, as presented in Table \ref{tab:main-performance}, demonstrate that our AttenHScore achieves significant performance improvements on both CoQA and SQuAD datasets. Specifically, our method outperforms other baseline methods across various evaluation metrics and exhibits stable improvements across different LMs. On TriviaQA and NQ datasets, we observe that the methods based on perplexity and AVG-Range exhibit larger variations in performance compared to their performance on CoQA and SQuAD. This is related to the fact that answers in the TriviaQA and NQ datasets are generally simpler and shorter. Our proposed method exhibits superior performance when handling complex questions. With respect to simpler questions, its performance is comparable to that of state-of-the-art methods.

\subsubsection{Collaborative Performance of LLMs and SLMs in QA}
By integrating our proposed model hallucination discrimination method and re-ranking strategy into the large-small LMs collaboration system, we conduct further experiments on the MultiFieldQA-zh from the Longbench benchmark \cite{bai2023longbench}, with the specific setup detailed in Appendix \ref{sec:appendix2}. The results in Table \ref{tab:allllm} show that simply reordering the retrieved content before inputting it into SLMs achieves significant performance improvement of 3.37. This indicates that SLMs encounter information overload issues when processing lengthy contexts, and optimizing the semantic relevance of the input sequence can effectively alleviate the limitations of their attention mechanisms.

Under the condition of limiting the total number of LLMs calls to 40\%, we compare the impact of four real-time detection and calling methods on performance improvement and find that AttenHScore method performs more prominently in terms of enhancing performance. It is worth noting that in the four columns, we find the performance of model collaboration to be slightly better than using the LLM alone. This finding is consistent with the observation results presented in Figure \ref{fig:leidatu}. It also indicates that when dealing with certain RAG problems, the performance of SLMs is comparable to or even better than that of LLMs.

\begin{figure}[t]
    \centering
    \includegraphics[width=0.47\textwidth]{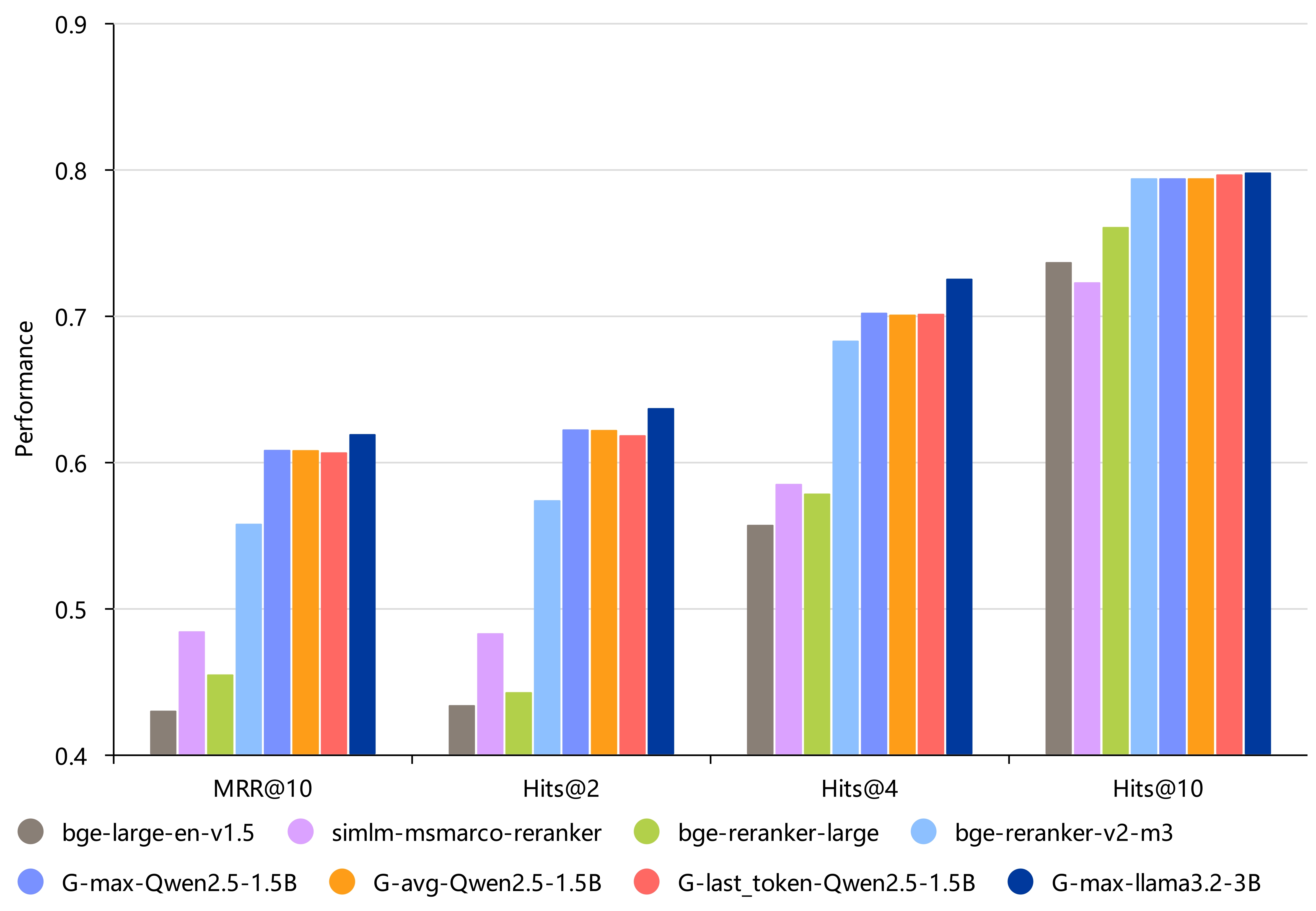}
    \caption{Performance comparison between the re-ranking method based on uncertainty evaluation and commonly used re-ranking models. Among them, the one starting with $G$ represents our approach, and the rest of the models are all from huggingface.}
    \label{fig:chongpai}
\end{figure}

\subsection{Comparison and Reflection on Re-ranking}
\label{Re-ranking}
In long text scenarios, SLMs struggle with information extraction and show a position bias, thus we introduce auxiliary mechanisms to enhance their capabilities. As shown in Figure \ref{fig:chongpai}, after retrieving the top-15 text chunks, we evaluate the relevance of the rearranged top-10 content. By comparing our proposed re-ranking method based on uncertainty $G$ with four existing re-ranking models, experimental results clearly demonstrate the excellent performance of our approach on the MRR@10, Hits@2, and Hits@4 metrics, indicating that our uncertainty can fully utilize the reasoning capabilities of LMs to more accurately identify texts relevant to the question. On the Hits@10 metric, our method slightly outperforms the most advanced re-ranking model, which is due to the incomplete retrieval results of top 15. In addition, we find that there is little difference in performance between using max, avg, and last-token to calculate attention scores, with max performing slightly better. Meanwhile, stronger LMs assisting uncertainty can further improve the performance of rearrangement.

\subsection{Ablation studies}
\label{Ablation studies}
The calculation methods of attention scores exhibit diversity, and we specifically test three methods listed in Table \ref{tab:2}. Experimental results reveal that the performance achieved using the max method surpasses that of the last-token and avg methods in both types of LMs. This superiority is primarily attributed to the fact that the max method is more effective in capturing the most prominent and critical information within the text sequence. In contrast, the last-token method tends to overly focus on the tail information of the sequence while neglecting other important elements, and the avg method tends to dilute the significance of key information due to averaging processing. This finding aligns with our proposed approach of detecting from the perspective of hallucination accumulation and transmission.

\renewcommand{\arraystretch}{1.2} 
\setlength{\extrarowheight}{1pt} 
\begin{table}[h]
\centering
\resizebox{0.47\textwidth}{!}{%
\begin{tabular}{lccc|ccc}
\toprule
\multicolumn{1}{l}{\textbf{Dataset}} & \multicolumn{3}{c}{\textbf{CoQA}} & \multicolumn{3}{c}{\textbf{SQuAD}} \\
\textbf{Attention} & \textbf{AUCs} & \textbf{ACCr} & \textbf{AUCr} & \textbf{AUCs} & \textbf{ACCr} & \textbf{AUCr}  \\
\midrule
\multicolumn{7}{c}{\textit{Llama3-8B-Instruct}} \\ 
\addlinespace[2pt] % Add space after italicized line
\cdashline{1-7}
last-token & 0.8226 & 0.8564 & 0.7948 & 0.8580 & 0.8864 & 0.8050  \\
avg & 0.8308 & 0.8673 & 0.8065 & 0.8678 & 0.8980 & 0.8176  \\
max & \textbf{0.8330} & \textbf{0.8706} & \textbf{0.8097} & \textbf{0.8715} & \textbf{0.9024} & \textbf{0.8176}  \\
\midrule
\multicolumn{7}{c}{\textit{Vicuna1.5-7B}} \\ 
\addlinespace[2pt] % Add space after italicized line
\cdashline{1-7}
last-token & 0.7473 & 0.8412 & 0.7675 & 0.7176 & 0.8014 & \textbf{0.7279}  \\
avg & 0.7491 & 0.8454 & 0.7792 & 0.7190 & 0.8059 & 0.7181  \\
max & \textbf{0.7503} & \textbf{0.8481} & \textbf{0.7840} & \textbf{0.7193} & \textbf{0.8085} & \underline{0.7212}  \\
\bottomrule
\end{tabular}%
}
\caption{Analysis of differences in three attention score calculation methods under different models.}
\label{tab:2}
\end{table}
%\multicolumn{2}{c}{\textbf{Dataset}}

\renewcommand{\arraystretch}{1.2} 
\setlength{\extrarowheight}{1pt} 
\begin{table}[h]
\centering
\resizebox{0.47\textwidth}{!}{%
\begin{tabular}{lccc|ccc}
\toprule
\multicolumn{1}{l}{\textbf{AUCs}} & \multicolumn{3}{c}{\textbf{SentenceSimilarity}} & \multicolumn{3}{c}{\textbf{Rouge-L}} \\
 \textbf{Method} & \textbf{0.7} & \textbf{0.8} & \textbf{0.9} & \textbf{0.3} & \textbf{0.5} & \textbf{0.7}  \\
\midrule
\multicolumn{7}{c}{\textit{Llama3-8B-Instruct}} \\ 
\addlinespace[2pt] % Add space after italicized line
\cdashline{1-7}
Perplexity & 0.5178 & 0.4898 & 0.4745 & 0.5528 & 0.5078 & 0.4937  \\
Energy & 0.4702 & 0.4423 & 0.4297 & 0.4885 & 0.4462 & 0.4333  \\
AVG-Range & 0.5016 & 0.4749 & 0.4609 & 0.5369 & 0.4957 & 0.4819  \\
LN-Entropy & 0.6185 & 0.6087 & 0.6113 & 0.6490 & 0.6288 & 0.6231  \\
LexicalSimilarity & 0.6549 & 0.6442 & 0.6365 & 0.6821 & 0.6640 & 0.6507  \\
EigenScore & 0.7303 & 0.7327 & 0.7359 & 0.7433 & 0.7397 & 0.7381  \\
AttenHScore & \textbf{0.8207} & \textbf{0.8498} & \textbf{0.8715} & \textbf{0.8373} & \textbf{0.8618} & \textbf{0.8733}  \\
\bottomrule
\end{tabular}%
}
\caption{Impact of correctness thresholds on hallucination detection performance.}
\label{tab:3}
\end{table}

\subsection{Hyper-parameter Sensitivity Analysis}
Utilizing the Llama3-8B-Instruct model, we execute comprehensive ablation experiments on the SQuAD dataset. The experimental results, shown in Table \ref{tab:3}, clearly demonstrate that different thresholds for correctness metrics have a significant impact on the final performance of hallucination detection. More importantly, our proposed AttenHScore exhibits superior performance compared to other baseline methods across various threshold settings.

On the other hand, we also carry out experiments on the decoding sampling hyperparameters of LMs, with specific results presented in Figures \ref{fig:temperture} and \ref{fig:top_k}. Experimental data reveals that our approach shows remarkable robustness across a wide range of parameter configurations.

Furthermore, considering the variability in the length of answers generated by SLMs, we introduce a preset token count $K$ during the calculation of hallucinatio, as specifically illustrated in Figures \ref{fig:K_value1} and \ref{fig:K_value2}. Our approach involves calculating an AttenHScore value for every $K$ tokens, and then selecting the maximum AttenHScore computed from the entire answer generated by SLMs as the basis for evaluation. Through observation, we find that system performance reaches an optimum when $K$ is set between 10 and 20. Further details of the experimental design and analysis are provided in Appendix \ref{sec:appendix3}.

\section{Conclusion}
Amidst the drive for efficiency and resource optimization, this study delves into the challenges of hallucination detection and prompt re-ranking within the collaboration of large and small LMs. We introduce a novel invocation discriminant metric, AttenHScore, which quantifies the accumulation and propagation of hallucinations in SLMs generations, enabling more precise detection of potential reasoning errors. Additionally, within a retrieval-based QA context, we steer SLMs to assess the uncertainty of queries relative to various text chunks, thereby achieving superior re-ranking and enhanced accuracy. Extensive experiments across four datasets reveal that our proposed real-time, plug-and-play detection methodology and re-ranking strategy strike an effective balance between cost and performance, eliminating the need for domain-specific knowledge or model training. We anticipate that our insights will inspire further researches into hallucination detection and re-ranking, ultimately promoting the development of collaboration between large and small LMs.

\section*{Limitations}
We acknowledge certain limitations, particularly in relying on the internal states of the LLM for hallucination detection. While this approach can identify hallucinations to some extent, there is still room for improvement in its accuracy. Future work will focus on deeper exploration of the LLMs' internal states to further enhance the precision and reliability of hallucination detection. Additionally, despite demonstrating good performance in complex query tasks, there may still be deficiencies in handling extremely complex tasks or those requiring deep semantic understanding. For instance, tasks involving multi-hop reasoning or strong domain relevance may not be fully addressed by the current invocation strategy. The primary objective of this paper is to further enhance the performance of the current large-small LM collaboration system through more accurate hallucination detection techniques. We will next concentrate on overcoming the limitations of existing methods to achieve a more efficient and reliable collaboration system.

\bibliography{custom}

% \newpage

\appendix
\section{Appendix}
\subsection{Hallucination Detection and Uncertainty Evaluation}
\label{sec:appendix1}
The concept of "hallucination" originally stems from the research domains of pathology and psychology, where it is defined as the perception of entities or events that do not exist in reality \cite{macpherson2013hallucination}. In the field of natural language processing (NLP), hallucination typically manifests as the generation that appears nonsensical or contradicts the original content \cite{maynez2020faithfulness}. Broadly speaking, hallucinations arising in NLP tasks can be classified into two major categories: intrinsic hallucination and extrinsic hallucination \cite{li2022faithfulness,ji2023survey}. The former refers to the conflict between the output content of LLMs and the original input information, while the latter refers to the generated content that cannot be verified by the original content.

As LLMs become increasingly adept at generating human-like text, distinguishing between accurate and hallucinated content has become a critical issue. Current research on hallucination detection requires access to the model's output content, latent states, or distributional features, and uncertainty assessment strategies based on the latter two have become an important research direction.

Fadeeva et al. \shortcite{fadeeva2024fact} introduce token-level and claim-conditioned uncertainty for fact-checking and entity -level detection. Varshney et al. \shortcite{varshney2023stitch} detect hallucinations by identifying tokens with low confidence, utilizing an active detection and mitigation pipeline. The analysis of Snyder et al. \shortcite{snyder2024early} involves examining softmax output probabilities, attention mechanisms, and gradients to identify early signs of hallucinations. The following approaches estimate uncertainty regarding meaning, rather than surface form, by considering entropy or semantic similarity over output distributions or samples. Semantic entropy \cite{farquhar2024detecting}, representing uncertainty at the meaning level, is introduced to robustly detect confabulations. MARS \cite{bakman2024mars}, a method that weights tokens based on semantic context in uncertainty scoring, is employed. Nikitin et al. \shortcite{nikitin2024kernel} propose a semantic similarity-based uncertainty quantification method for LLMs, where kernel language entropy is exploited to assess uncertainty via von Neumann entropy over semantically-clustered model outputs. This field acknowledges high-certainty hallucinations and calibration as key unresolved challenges, pushing for a deeper introspective and semantics-based analysis. Our detection pipelines integrate probability features, content perception, and attention mechanisms to form a comprehensive signal. 

\subsection{Analysis of Real-Time Capability}
\label{sec:efficiency}
The calculation of AttenHScore is based on the attention weights and generation probabilities produced by the model itself during the generation process. This information is naturally generated during inference and requires no additional computation. We simply leverage this readily available information for judgment, and the process is nearly instantaneous, thus the method introduces no additional time delays. Furthermore, our method is significantly more efficient compared to approaches that necessitate model training or multiple generations.

We highlight the following advantages exhibited by our method: (1) Unsupervised: As an evaluation metric for invocation, AttenHScore can be directly calculated without relying on any detector training process, simplifying the evaluation workflow. (2) Real-time: Compared to current post-processing methods, AttenHScore, as a real-time invocation detection metric, ensures the efficient evaluation process. (3) Plug-and-play: Designed as a lightweight algorithm, AttenHScore can be easily integrated into any existing Transformer-based LMs.

\subsection{Detailed Experimental Setup for Reproducibility}
\label{sec:appendix2}
All language models utilized in this paper employ the chat or instruct versions where multiple versions exist, and are loaded in full precision (Float32). The vector database is constructed using Milvus, where the embedding model for English texts is bge-large-en-v1.5\footnote{\url{https://huggingface.co/BAAI/bge-large-en-v1.5}}, and bge-base-zh-v1.5\footnote{\url{https://huggingface.co/BAAI/bge-base-zh-v1.5}} for Chinese texts. To more effectively verify the effectiveness of the component designed for detecting small-model hallucinations in the collaborative system of large-small LMs, we utilize three SLMs of different types and sizes: Llama3-8B-Instruct\footnote{\url{https://huggingface.co/meta-llama/Meta-Llama-3-8B-Instruct}}, Vicuna1.5-7B\footnote{\url{https://huggingface.co/lmsys/vicuna-7b-v1.5}}, and Llama2-13B-Chat-HF\footnote{\url{https://huggingface.co/meta-llama/Llama-2-13b-chat-hf}}. The sentence embeddings of model generation and the ground truth answer are extracted by the nli-roberta-large model\footnote{\url{https://huggingface.co/sentence-transformers/nli-roberta-large}}.

In Table \ref{tab:allllm}, we employ nine different LLM interfaces to conduct large-small LM collaborative experiments with Vicuna1.5-7B. These interfaces are as follows: ERNIE-3.5\footnote{\url{https://console.bce.baidu.com/qianfan}}, Qwen-Plus\footnote{\url{https://bailian.console.aliyun.com/}}, Qwen-Turbo\textsuperscript{8}, Deepseek-v3\footnote{\url{https://platform.deepseek.com/}}, Qwen-72B\footnote{\url{https://huggingface.co/Qwen/Qwen-72B-Chat}}, Qwen1.5-72B\footnote{\url{https://huggingface.co/Qwen/Qwen1.5-72B-Chat}}, Qwen2-57B\footnote{\url{https://huggingface.co/Qwen/Qwen2-57B-A14B-Instruct}}, DeepSeek-R1-Distill-Llama-70B\footnote{\url{https://huggingface.co/deepseek-ai}}, and DeepSeek-R1-Distill-Qwen-32B\textsuperscript{13}. Our experimental setup involves retrieving 10 relevant documents for each query and having the SLM to generate responses accordingly. Subsequently, different hallucination detection methods are utilized to monitor the generation status of the SLM in real-time. If it is determined that the SLM's output contains hallucinations, the corresponding LLM interface is invoked to answer the question. Regarding text chunking operations, we adopt the LLM-based chunking method \cite{jihao2019meta}.

\subsection{Exploring Hyperparameter Settings for Optimal Performance}
\label{sec:appendix3}
Different hyperparameter settings may not only serve as critical factors influencing model performance, but also exert differential impacts on the sensitivity of various detection methods. Consequently, we conduct a systematic analysis of hyperparameters including temperature, top-k and $K$.

\begin{figure}[h]
    \centering
    \includegraphics[width=0.49\textwidth]{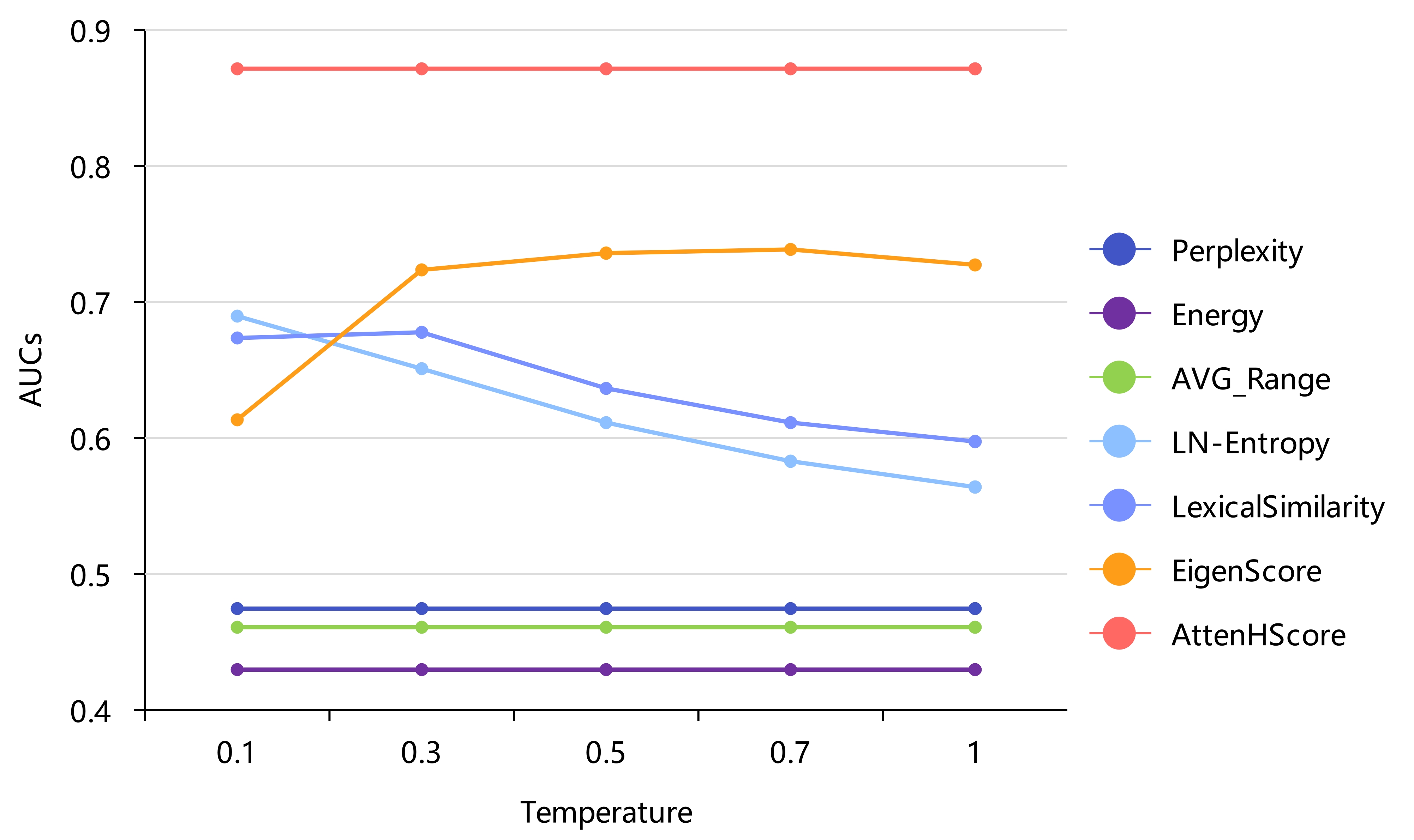}
    \caption{Performance sensitivity to temperature on Dataset SQuAD.}
    \label{fig:temperture}
\end{figure}

\begin{figure}[h]
    \centering
    \includegraphics[width=0.49\textwidth]{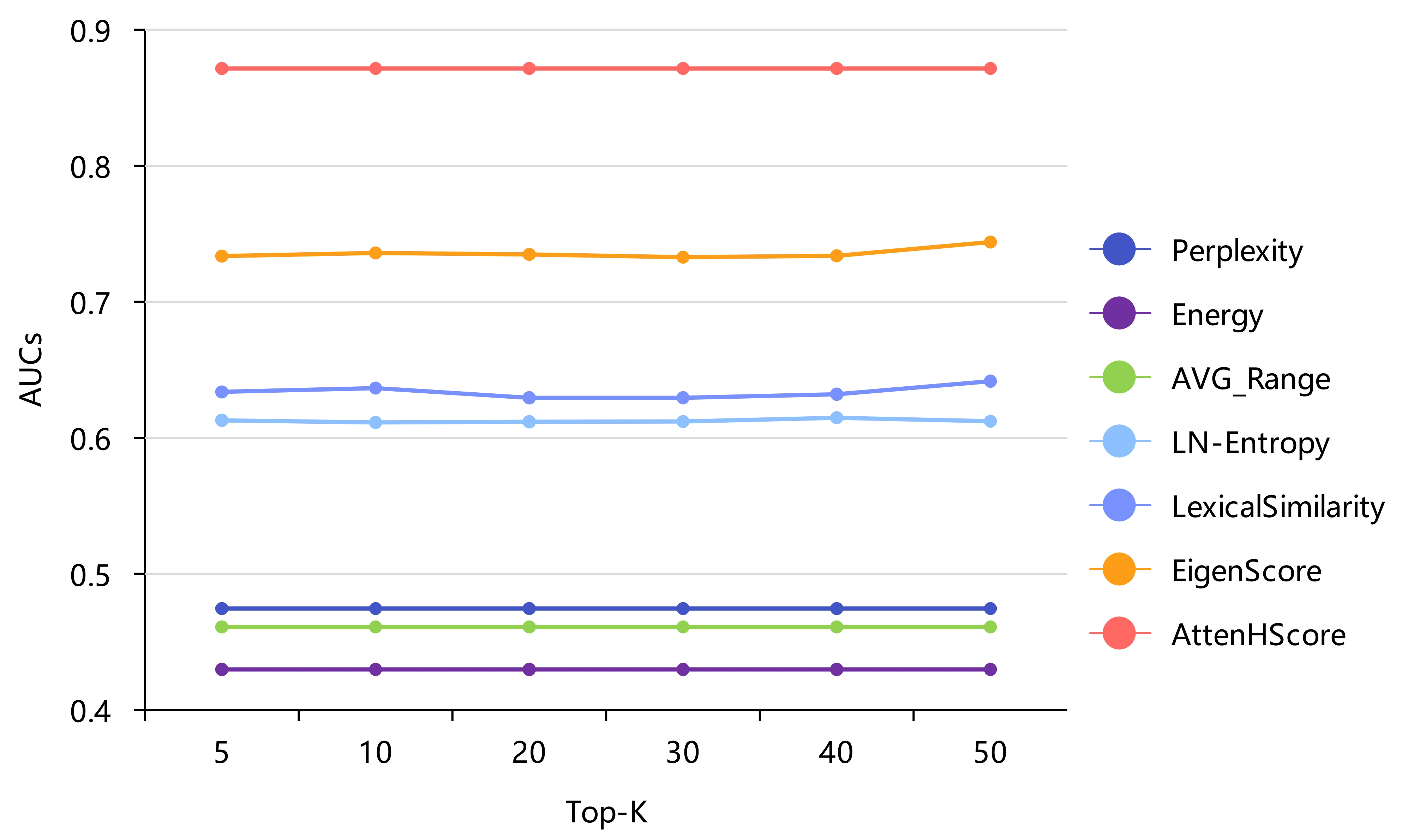}
    \caption{Performance sensitivity to top-k on Dataset SQuAD.}
    \label{fig:top_k}
\end{figure}

Experimental data reveals that various detection methods exhibit relatively low sensitivity to the top-k, whereas LN-Entropy, LexicalSimilarity, and EigenScore demonstrate higher sensitivity to the temperature. Extensive experiments in Figures \ref{fig:temperture} and \ref{fig:top_k} confirm that our approach shows remarkable robustness across a wide range of parameter configurations.

\begin{figure}[h]
    \centering
    \includegraphics[width=0.47\textwidth]{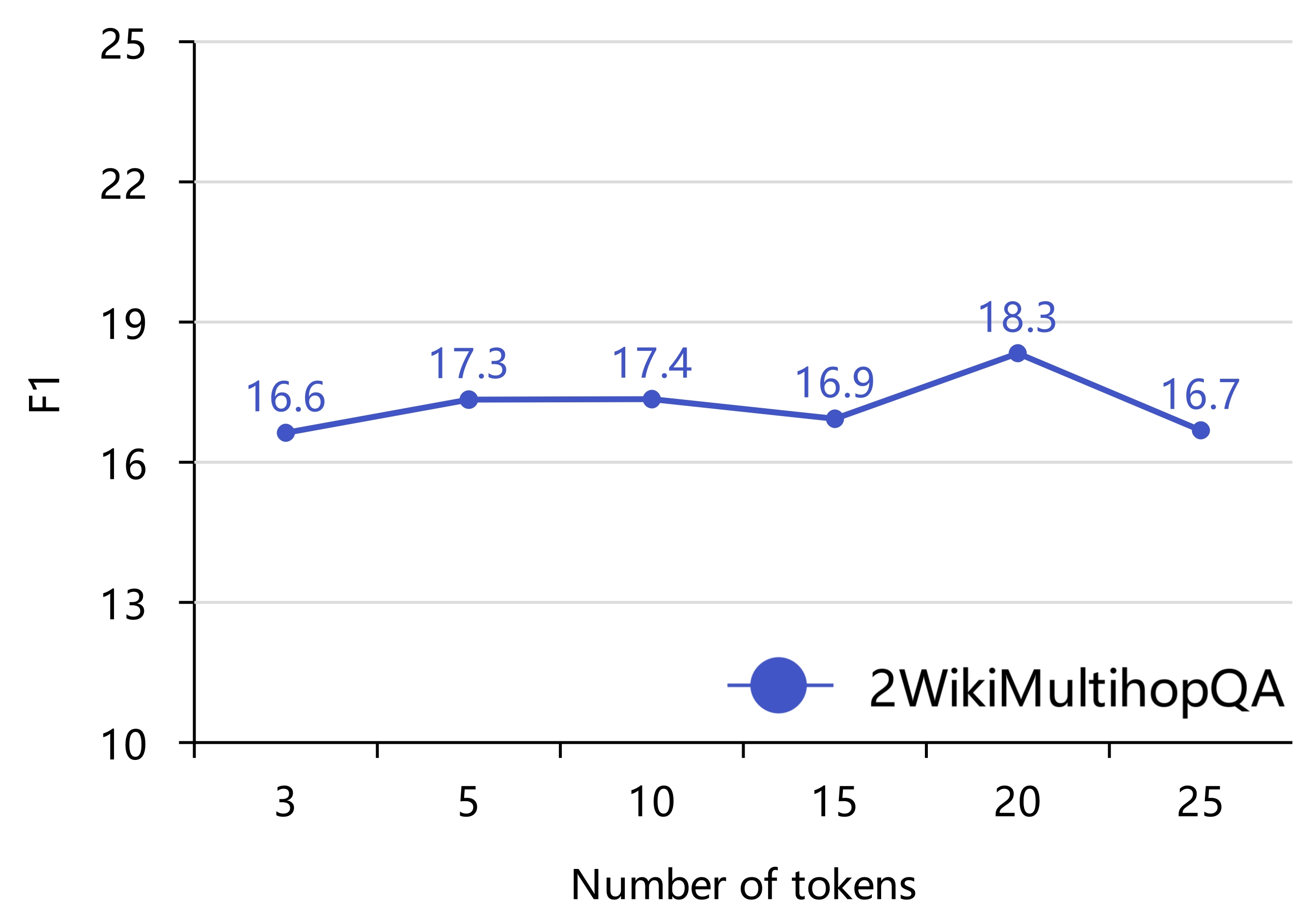}
    \caption{Performance sensitivity to $K$ (Number of tokens) on Dataset 2WikiMultihopQA.}
    \label{fig:K_value1}
\end{figure}

\begin{figure}[h]
    \centering
    \includegraphics[width=0.47\textwidth]{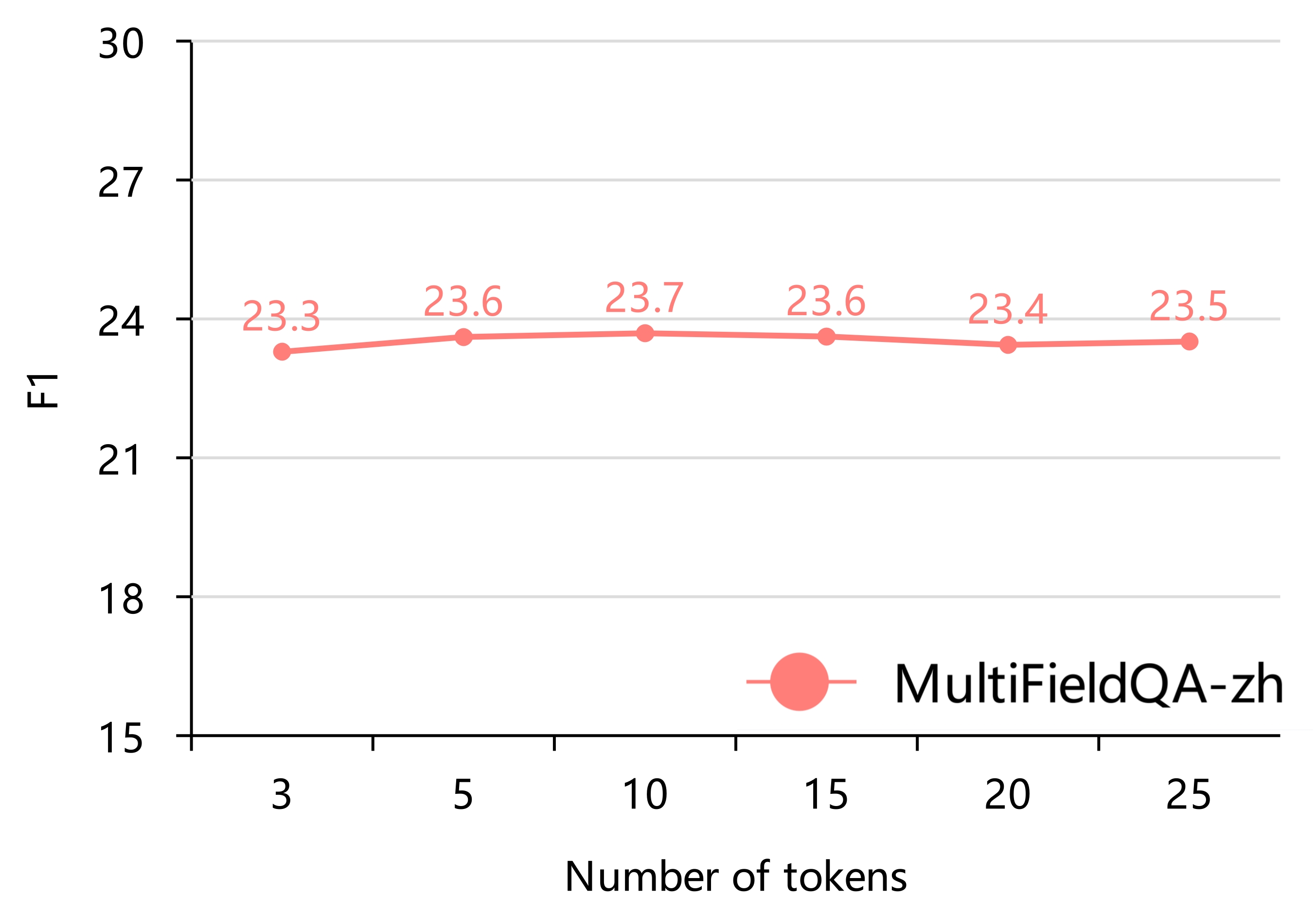}
    \caption{Performance sensitivity to $K$ (Number of tokens) on Dataset MultiFieldQA-zh.}
    \label{fig:K_value2}
\end{figure}

In the experimental section described in Figures \ref{fig:K_value1} and \ref{fig:K_value2}, we conduct a detailed comparative analysis of the performance across different values of $K$. The results indicate that the system achieves optimal performance when $K$ is set between 10 and 20 tokens.



\subsection{Setting Method for Dynamic Threshold}
\label{sec:threshold}
We adopt an adaptive strategy for threshold setting. Specifically, we first calculate the initial threshold using the average hallucination score of the first five queries. Subsequently, for each new query, we incorporate the current query's hallucination score into the historical records and recalculate the average hallucination score of all processed queries, using this as the updated threshold.
\[
    \theta  = \frac{\sum_{i=1}^{n} S_{RHDI}(X_i)}{n}
\]
     
In real-world production environments, systems are typically reused multiple times. We utilize the outputs from the first five queries to calculate the initial threshold. As each query is processed, the system records and dynamically computes the average hallucination score of previously generated answers in real time, thereby continuously adjusting the threshold. The update mechanism of dynamic threshold is independent upon the dataset.

For a strong model, which tends to produce lower average scores, its dynamic threshold will naturally be lower as well. Only when a question's score significantly exceeds this low baseline is it classified as a hallucination. For a weak model, which generates higher average scores, its dynamic threshold will be correspondingly higher. The system determines hallucations based on whether the current score significantly surpasses this high baseline.

Our dynamic threshold is not a fair yardstick aimed at measuring the absolute difficulty of all problems. Instead, it is an adaptive benchmark that reflects the typical performance of the current model when handling a recent stream of tasks. If the system encounters a series of complex problems in the initial stage, resulting in a relatively high average score, then a higher threshold is precisely reasonable as it accurately reflects the model's current operational state. The system's goal is to determine whether the score of the current problem is abnormal relative to the model's own average performance, rather than based on its absolute value.

The threshold is calculated based on the rolling average of all processed problems. This means that the influence of any early, occasional extreme high scores will be continuously diluted as the number of processed problems increases. In real-world scenarios where the system operates over an extended period , the bias introduced by the initial few problems will become negligible.

\begin{figure*}[h!]
    \centering
    \includegraphics[width=\textwidth]{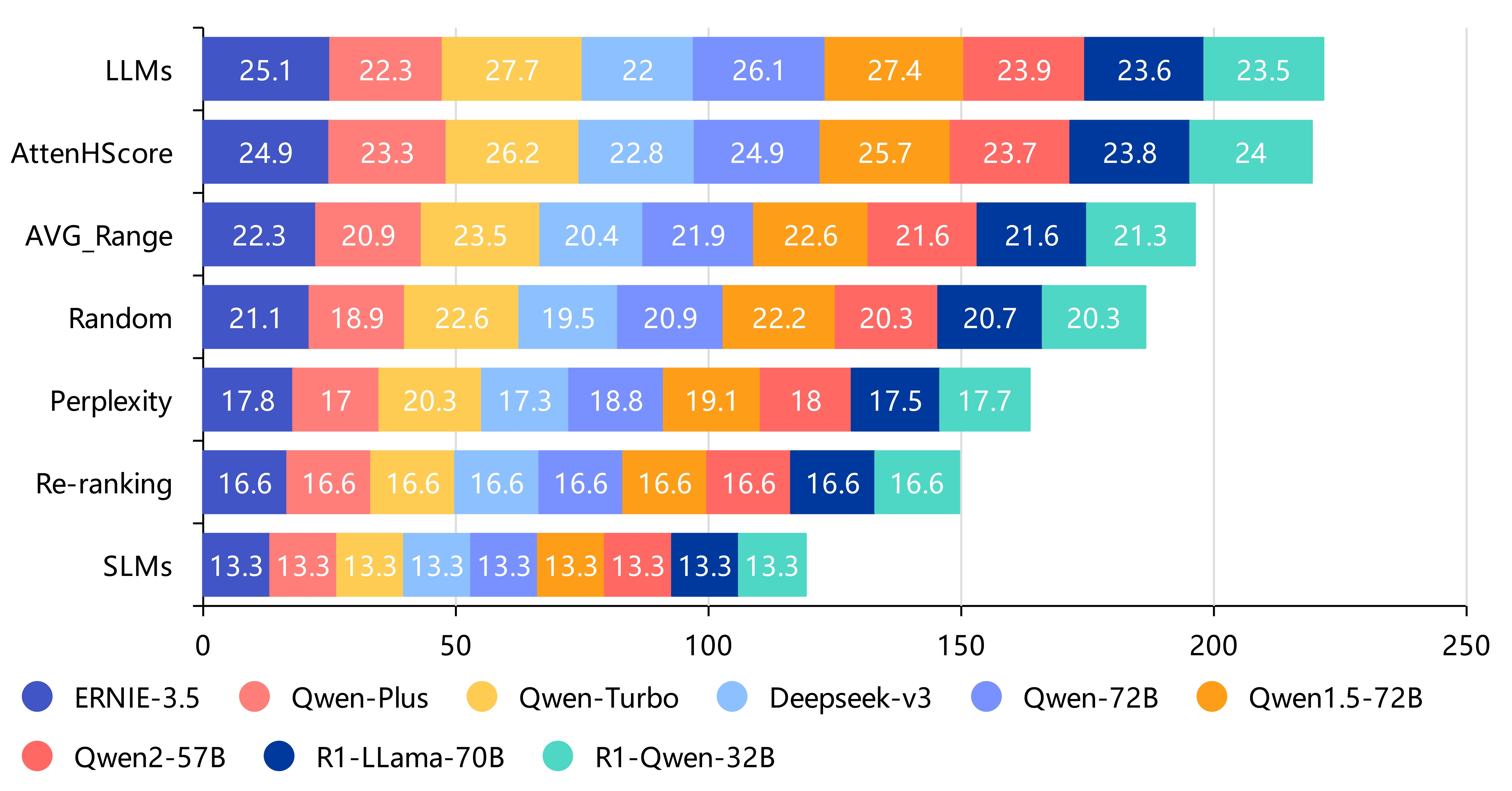}
    \caption{Comparative snalysis of AttenHScore and other methods in large-small LM collaboration system.}
    \label{fig:xietong}
\end{figure*}

\begin{figure*}[h!]
    \centering    
    \includegraphics[width=\textwidth]{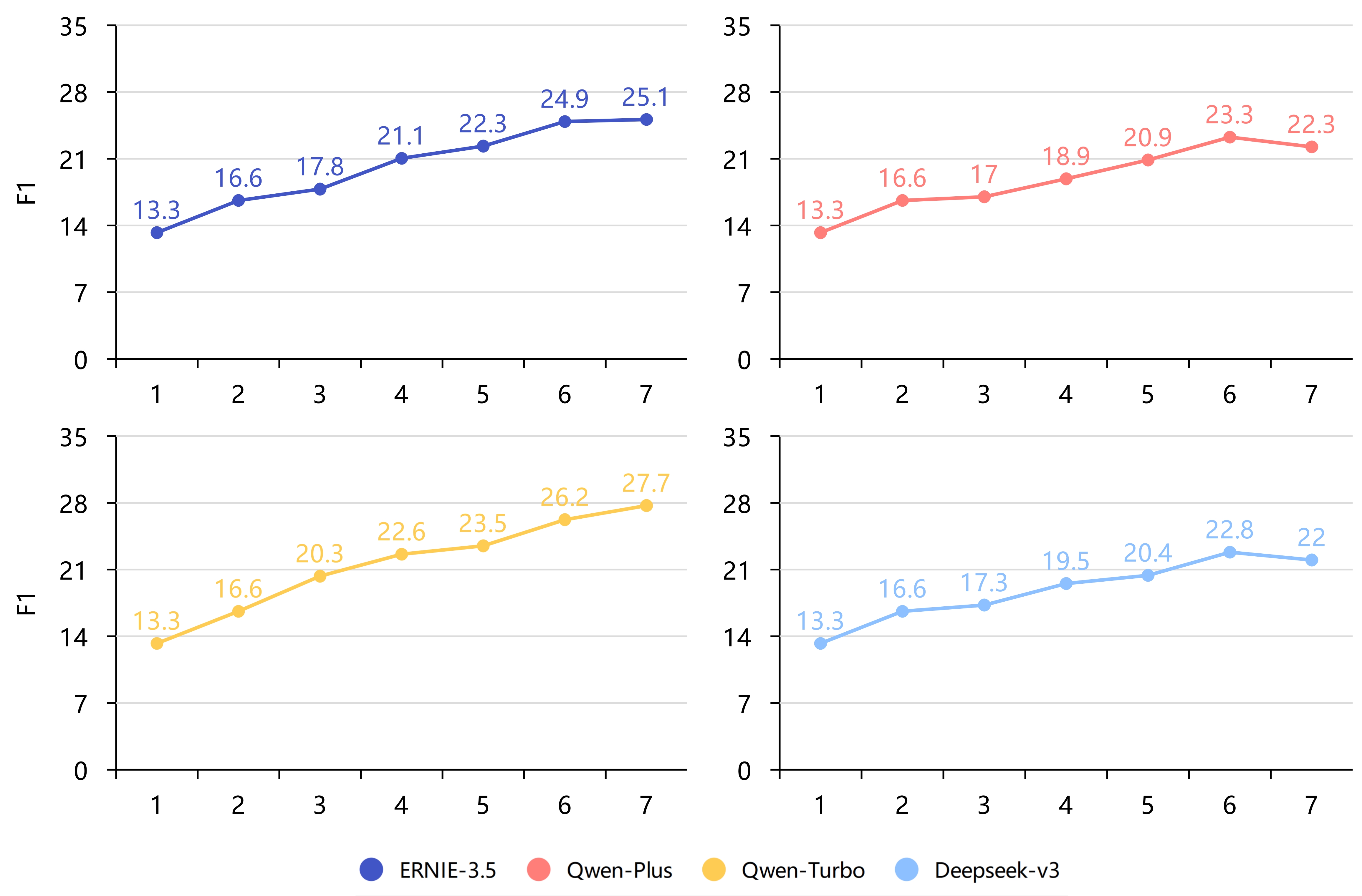}
    \caption{Performance variation trends of various large-small LM collaboration methods Under different LLM interfaces. The approaches include: 1: SLMs, 2: Re-ranking, 3: Perplexity, 4: Random, 5: AVG\_Range, 6: AttenHScore and 7: LLMs.}
    \label{fig:xietong2}
\end{figure*}

\subsection{Further Exploration of Large-Small LM Collaboration}
\label{sec:appendix4}
We conduct a more in-depth analysis and visualization of the experiments on the collaboration between large and small LLMs presented in Table \ref{tab:allllm}. As shown in Figure \ref{fig:xietong}, we accumulate the performance of SLMs, re-ranking, four real-time collaboration strategies, and LLMs, where each color represents the performance of a method under the corresponding LM interface. The scores of LLMs called separately and the collaboration system using AttenHScore as the hallucination detection component are relatively similar, indicating that our metric is more effective in identifying hallucinated information generated by SLMs. In Figure \ref{fig:xietong2}, we also demonstrate the performance trends of different methods under some LLMs through line charts. It can be observed that the overall data displays an upward trend, and two charts even have higher points at AttenHScore than when using only the LLM, which more directly illustrates the superiority of our method.

\end{document}